\documentclass[journal]{IEEEtran}
\ifCLASSINFOpdf
\else
\fi
\usepackage[utf8]{inputenc} 
\usepackage[T1]{fontenc}    
\usepackage[bookmarks=false,colorlinks]{hyperref}       
\usepackage{url}            
\usepackage{booktabs}       
\usepackage{amsfonts}       
\usepackage{nicefrac}       
\usepackage{microtype}      
\usepackage{xcolor}         
\usepackage{orcidlink}
\usepackage{times}
\usepackage{helvet}
\usepackage{courier}
\usepackage{amsmath,amssymb}
\usepackage{algorithmic}
\usepackage{algorithm}
\usepackage{array}
\usepackage{textcomp}
\usepackage{stfloats}
\usepackage{url}
\usepackage{verbatim}
\usepackage{graphicx}
\usepackage{epsfig}
\usepackage{booktabs}
\usepackage{makecell}
\usepackage{multirow}
\usepackage{pifont}
\newcommand{\xmark}{\ding{55}}%
\usepackage{tablefootnote}

\usepackage{wrapfig}
\usepackage{enumitem}

\DeclareMathOperator\arctanh{arctanh}
\usepackage{subcaption}
\begin{document}
%
\title{Sharpness-Aware Geometric Defense for Robust Out-Of-Distribution Detection}
%
%
%

\author{Jeng-Lin Li,~\IEEEmembership{Member,~IEEE,}
        Ming-Ching Chang,~\IEEEmembership{Senior Member,~IEEE,}
        and~Wei-Chao Chen,~\IEEEmembership{Member,~IEEE}
\thanks{JL. Li and WC. Chen are with Inventec Corporation.}
\thanks{MC. Chang is with the University at Albany -- State University of New York.}
\thanks{Manuscript received February 26, 2025.}}

%
%

\markboth{Journal of \LaTeX\ Class Files,~Vol.~14, No.~8, August~2015}%
{Shell \MakeLowercase{\textit{et al.}}: Bare Demo of IEEEtran.cls for IEEE Journals}
%



\maketitle

\begin{abstract}
  Out-of-distribution (OOD) detection ensures safe and reliable model deployment. Contemporary OOD algorithms using geometry projection can detect OOD or adversarial samples from clean in-distribution (ID) samples. However, this setting regards adversarial ID samples as OOD, leading to incorrect OOD predictions. Existing efforts on OOD detection with ID and OOD data under attacks are minimal. In this paper, we develop a robust OOD detection method that distinguishes adversarial ID samples from OOD ones. The sharp loss landscape created by adversarial training hinders model convergence, impacting the latent embedding quality for OOD score calculation. Therefore, we introduce a {\bf Sharpness-aware Geometric Defense (SaGD)} framework to smooth out the rugged adversarial loss landscape in the projected latent geometry. Enhanced geometric embedding convergence enables accurate ID data characterization, benefiting OOD detection against adversarial attacks. We use Jitter-based perturbation in adversarial training to extend the defense ability against unseen attacks. Our SaGD framework significantly improves FPR and AUC over the state-of-the-art defense approaches in differentiating CIFAR-100 from six other OOD datasets under various attacks. We further examine the effects of perturbations at various adversarial training levels, revealing the relationship between the sharp loss landscape and adversarial OOD detection. 
\end{abstract}


%
\IEEEpeerreviewmaketitle

\section{Introduction}
\label{sec:intro}
\vspace{-1mm}

In multimedia, {\em out-of-distribution (OOD)} detection~\cite{shen2021towards,yang2021generalized} is crucial for ensuring AI models can differentiate between familiar and unfamiliar content, enhancing reliability in real-world applications. Multimedia AI systems define their in-distribution (ID) space yet these models encounter unseen content such as an image classifier trained on urban scenes but tested on underwater footage. Effective OOD detection enables models to recognize when a sample deviates from the training distribution, prompting actions like rejecting uncertain inputs, requesting human verification, or adapting to new data. 
Despite their importance, the robustness of these OOD systems remains insufficiently explored. They are particularly vulnerable to {\em adversarial attacks}~\cite{goodfellow2015explaining} and frequently struggle to detect irregular or unanticipated classes when confronted with real-world adversaries.

\begin{figure}[t!p]
\centerline{\includegraphics[width=\linewidth]{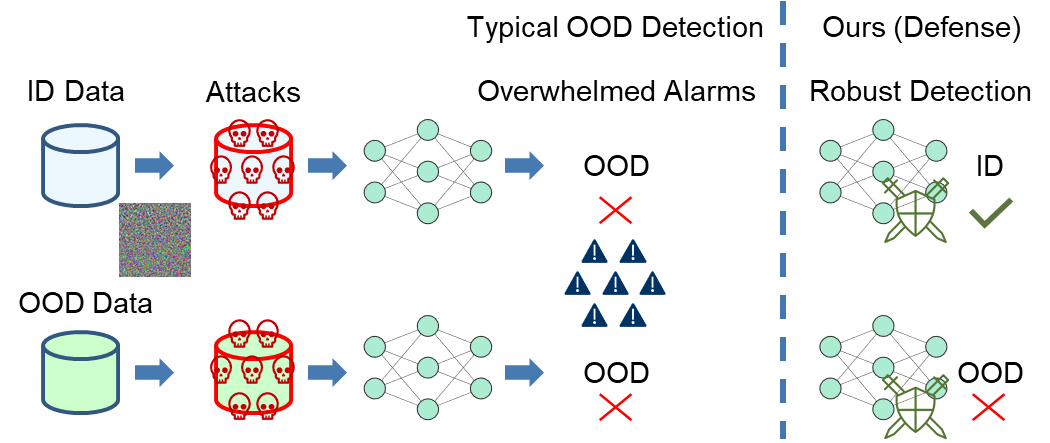} 
}
\caption{Adversarial attacks lead to the failure of OOD detection, leading to unsafe model deployment.}
\label{fig:teaser}
\end{figure}

The OOD research field primarily focuses on detecting unseen classes but often mishandles the risks posed by frequent real-world adversaries. Past OOD detection studies predict adversarial samples as OOD samples~\cite{lee2018simple}, unnecessarily leading to substantial alarms for adversarial ID cases shown in Figure~\ref{fig:teaser}. As a result, OOD systems require further enhancement by incorporating insights from adversarial attack research. These systems can be fortified by leveraging defense methods to remain robust against subtle yet damaging manipulations that disrupt OOD predictions. 
A handful of adversarial defense studies are proposed to secure the model classification against the attacks~\cite{rade2022reducing,pang2020boosting,fakorede2023improving}. Notably, {\em adversarial training} and {\em hyperspherical geometry learning} effectively alleviate adversarial situations in the image classification tasks. Inspired by these adversarial defense studies, we aim to ensure the OOD detection system resiliently operates in both clean and adversarial conditions. 
The task of differentiating OOD itself is hard due to the widespread new data pattern to the model~\cite{fang2022out}, and suffering from adversarial attacks increases the complexity of OOD detection. ATOM is a pioneering framework for dealing with attacks on open-set samples~\cite{chen2021atom}. Recently, Azizmalayeri {\em et al.}~\cite{azizmalayeri2022your} found that adversarial attacks on both ID and OOD data significantly degrade detection accuracy. They introduced an Adversarial Training Discriminator (ATD) with an outlier exposure strategy that simulates both adversarial ID and OOD samples. The outlier exposure method highly depends on the auxiliary OOD datasets which are expected to be excessively large. This requirement leads to inefficient and impracticality in real-world applications. We target the defense against challenging {\em white-box} attacks on both ID and OOD data and seek effective perturbation strategies without relying on additional large outlier datasets.

In this work, we aim to tackle this robust OOD detection issue by combining two perspectives, including \textit{geometry optimization} and \textit{loss landscape smoothing}. 
First, the hypersphere and hyperbolic geometries can learn compact representations for OOD detection~\cite{ming2023how}, but we still empirically observe a high false positive rate. Therefore, we further examine the ability of the multiple-geometric learning method~\cite{li2024learning} in accommodating ID data variability under adversarial attacks.
Second, sharp loss has been observed in prior research~\cite{losslandscape,adv_loss} which is caused by adversarial samples. These samples increase the gradient norm and the subsequent local minimum, sharp loss, and challenges in convergence. The GAN-like structure in ATD is also known for its loss convergence issues. 

Therefore, we introduce a sharpness-aware adversarial training framework that effectively alleviates the sharp loss landscape, achieving robust latent geometry learning. Our backbone network learns a Multi-Geometry Projection (MGP)~\cite{li2024learning} by incorporating two Riemannian (hypersphere and hyperbolic) geometries with distinct curvatures to fully characterize the complex ID data. 
In the adversarial training procedure, we propose to utilize the Riemannian Sharpness-aware Minimization (RSAM)~\cite{yun2023riemannian,truong2023rsam}, which improves the multiple Riemannian geometry convergence by flattening the adversarial loss landscape. 
We empirically find that performing adversarial training based on the \textit{Jitter} attacks~\cite{schwinn2023exploring} demonstrates generalizability in defending against other attacks.

Our experiments comprehensively investigate mainstream OOD detection approaches with and without adversarial training. We use CIFAR-10 and CIFAR-100 as ID datasets and perform OOD evaluation using six other datasets. 
We compare the proposed SaGD against ATD~\cite{azizmalayeri2022your}, the state-of-the-art (SoTA) defense approach for OOD detection, and show our improvements. Additionally, we examine the effects of different adversarial training approaches to reveal the generalization ability of SaGD in defending other types of attacks. 
In contrast to other OOD studies~\cite{azizmalayeri2022your,chen2021atom,chen2021robust}, which only present one type of Projected Gradient Descent (PGD) attack, our results are comprehensively from the average of six conditions, including the case without attack and five other cases under different attacks. We report the area under the ROC curve (AUC) along with the false positive rate at 95\% true positive rate (abbreviated as FPR$_{95}$) as evaluation metrics. The FPR$_{95}$ is a common metric for OOD detection; however, it is not reported in \cite{azizmalayeri2022your}. 
In the adversarial OOD detection experiments using the CIFAR-10 ID dataset, our SaGD robustly reduces 14.91\% FPR$_{95}$ and enhances 7.47\%  AUC over the SoTA approach. We also achieved a 17.71\% average FPR$_{95}$ reduction and 10.18\% AUC improvement using CIFAR-100 as the ID dataset.


Our contribution is summarized as follows:
%
\begin{itemize}[leftmargin=10pt] \itemsep -.1em

\item We introduce a novel sharpness-aware method for improving OOD detection in adversarial training. Our method investigates the combination of Riemannian geometries under adversarial conditions. 
This expansion of geometry space sharpens our defense against adversarial attacks and avoids reliance on large OOD datasets for auxiliary training.

    
\item We examine different perturbation techniques (not limited to PGD) for adversarial training to identify their effectiveness for robust OOD detection.
    
\item We investigate various adversarial attacks on different OOD detection approaches and report results on FPR$_{95}$ and AUC. Our SaGD sets a new SoTA for OOD detection, excelling in FPR$_{95}$ and AUC metrics, both with or without attacks.
    
\item We analyze the relations between the minimization of a sharp loss landscape and OOD detection performance under various adversarial conditions.

\end{itemize}

\begin{figure*}[t]
\centerline{
  \includegraphics[width=\linewidth]{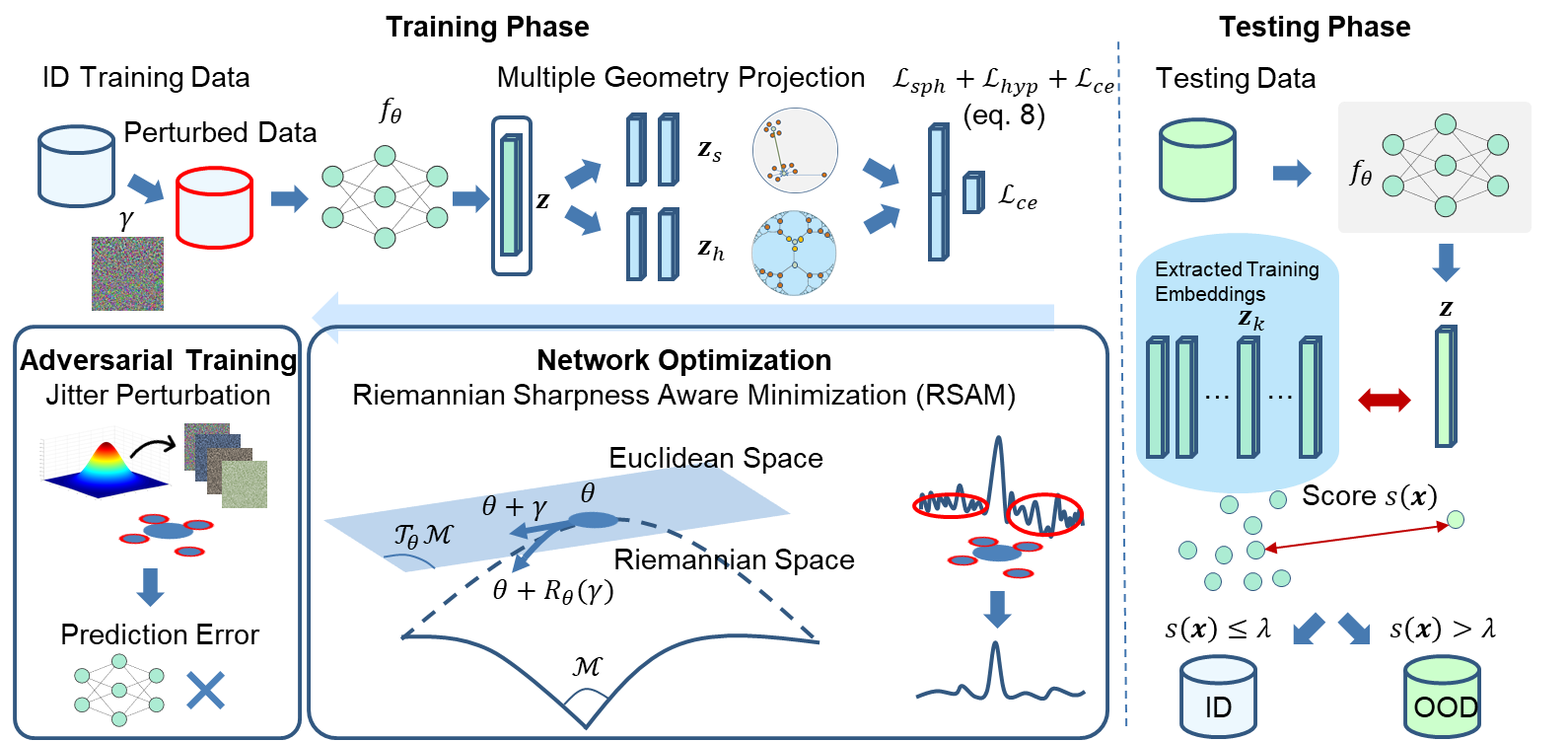} 
}
\caption{Overview of the proposed \textit{Sharpness-aware Geometry Defense (SaGD)} framework for robust OOD detection. The Multi-Geometry Projection (MGP) network is trained using Jitter-based adversarial samples and optimized via sharpness-aware loss minimization using RSAM. In testing, sample embedding is computed for scoring to discern OOD from ID cases.}
\label{fig:framework}
\end{figure*}

\section{Related Work}
\label{sec:related_work}

\subsection{OOD Detection}


\noindent
{\bf Post-processing OOD detection.} Model-agnostic OOD detection methods~\cite{shen2021towards,yang2021generalized} formulate scoring functions based on prediction probability and energy score. 
Determining prediction confidence can take various forms, such as utilizing softmax outputs~\cite{hendrycks2016baseline}, energy-based scores~\cite{liu2020energy}, or entropy functions~\cite{chan2021entropy}. 
To avoid re-training or excessive tuning of the given model, recent advancements focus on introducing perturbation~\cite{liang2018enhancing}, conducting pruning~\cite{djurisic2022extremely}, and generating an unknown novel class~\cite{wang2022vim} to enhance the distinction between OOD and in-distribution (ID) samples. 

\noindent
{\bf Model training OOD detection.}
Other OOD studies have sought to enhance fixed-model post-processing by incorporating network constraints during training to improve OOD detection. Sophisticated designs are devised for network space projection and embedding distance measurement to effectively train models for OOD detection. Noteworthy examples include SSD~\cite{sehwag2020ssd} and KNN+\cite{sun2022out}, employing contrastive loss for latent embedding learning and calculating Mahalanobis\cite{lee2018simple} and non-parametric KNN distances, respectively. A recent addition to this line of work is the CIDER framework~\cite{ming2023how}, which has demonstrated improved OOD detection performance by imposing constraints on samples using a hypersphere-based loss function. The hyperbolic embedding also demonstrates the enhanced ability for OOD detection~\cite{Guo_2022_CVPR}. 
Despite the impressive results achieved by these OOD detection approaches, their performance is not robust when facing adversarial samples in practice.

\subsection{Adversarial Defenses}


Adversarial training~\cite{madry2018towards} stands out as a key defense against adversarial attacks. This method involves integrating adversarial samples into model training to bolster the network's resilience against perturbations. The goal is to approximate potential perturbations in adversarial samples, using them to enhance model accuracy. However, the incorporation of adversarial samples unavoidably leads to a degradation in prediction accuracy due to the introduction of various noises.
To this end, Helper-based Adversarial Training (HAT) seeks balance and reduces harm from adversarial samples by tailoring network architecture and loss designs~\cite{rade2022reducing}. Notably, works such as~\cite{shao2020open, shao2022open} train dual-attentive denoising layers, leading to clean reconstructed samples from adversarial ones. Originally devised for addressing the open-set detection problem, these techniques find application in OOD detection scenarios under adversarial attacks~\cite{azizmalayeri2022your}.

Outlier exposure~\cite{hendrycks2018deep,pei2022out} emerges as a strategy in OOD detection, broadening its capabilities by incorporating outliers during training. Although these techniques can also foster a robust learning space for adversarial outliers~\cite{chen2021robust}, their practical utility is constrained by the uncertainty surrounding the optimal inclusion of outliers and the types of adversarial that should be artificially introduced.


\subsection{Regularization and Adversarial Robustness}

Regularizing neural networks proves effective against adversarial attacks by preventing the adoption of overly complex parameters, and avoiding suboptimal convergence at saddle points. When attacked samples have distributions vastly different from the training space, it significantly biases the network parameters. Implementing regularization based on various network designs, such as angular and margin regularization on hypersphere geometry, enhances adversarial robustness~\cite{pang2020boosting, fakorede2023improving}. Prioritizing more regularization for vulnerable samples minimizes the robustness risk and improves generalizability~\cite{yang23h}. The underlying reason is that adversarial training can generate rugged sample space and thus hinder model convergence. Such a sharp loss landscape hinders the training process with scattered gradients and increased curvatures~\cite{liu2020loss}.

Sharpness-Aware Minimization (SAM)~\cite{foret2021sharpnessaware} is a well-known technique for its regularization ability to mitigate training overfitting on a sharp loss landscape. 
A recent study~\cite{wei2023sharpness} delves into SAM's potential for adversarial robustness and empirically establishes a lightweight alternative to PGD adversarial training without significant sacrifices in clean sample accuracy.
However, the integration of SAM-based regularization with adversarial training, especially in OOD detection, remains limited. The exploration of geometric projection associations, such as RSAM operating on the Riemannian manifold~\cite{yun2023riemannian, truong2023rsam}, is largely uncharted.
This paper advances current research by combining RSAM with multi-geometry learning techniques for OOD detection and exploring the adversarial conditions. 


\section{Sharpness-aware Geometry Defense (SaGD)}
\label{sec:method}

Figure~\ref{fig:framework} overviews our proposed SaGD framework, where the training phase consists of adversarial training using Jitter-based adversarial samples ($\S$\ref{sssec:adversarial_training}), multi-geometry projection ($\S$\ref{sssec:mmel}), and sharpness-aware optimization ($\S$\ref{sssec:optimization}). 
The multi-geometry backbone combines the hypersphere and hyperbolic branches in a multi-task joint loss optimization scheme. We first introduce the architecture along with the scoring function for OOD score calculation. We then show how the Riemannian Sharpness-aware Minimization (RSAM) optimizes the framework with adversarial training. 

\noindent
{\bf Problem setup.}
Given labeled data $(x, y)$ from a distribution $\cal D$, we consider a model $f_\theta$ with parameters $\theta$. The training and testing data are denoted as ${\cal D}_s$ and ${\cal D}_t$, respectively, where ${\cal D}_t$ contains both in-distribution (${\cal D}_{id}$) and out-of-distribution (${\cal D}_{ood}$) test data.  We assume ${\cal D}_{id}$ is drawn from the same distribution as ${\cal D}_s$, while ${\cal D}_{ood}$ is from a different distribution that needs to be distinguished.
The standard procedure for OOD detection is as follows:
(1) Train a model $f_\theta$ with ${\cal D}_s$.
(2) Fix model parameter $\theta$ during test time. For each test sample $x$, derive embedding $z$ using $f_\theta$.
(3) Calculate OOD score $s(x)$ and differentiate OOD samples with a threshold $\lambda$.
To protect the model against adversarial attacks, we focus on the first step to strengthen the model's robustness. 

\subsection{Adversarial Training}
\label{sssec:adversarial_training}

We generate adversarial samples by utilizing the \textit{Jitter} adversarial attack~\cite{schwinn2023exploring}. 
Each input sample $x$ is perturbed by \textit{Jitter} attack to simulate the inference-time attacks. Denote the perturbed samples as $x_\gamma = x+\gamma$, where $||\gamma||_p \leq \epsilon$ with an $l_p$-norm bound and we select $p$ to be the infinite norm. 

The \textit{Jitter} attack rescales the softmax function as $\hat{\mathbf{h}} = softmax \left( \alpha \cdot \frac{\mathbf{h}}{||\mathbf{h}||_\infty} \right)$. This is based on a finding that a small value range of output logits $\mathbf{h}$ can reduce the attack success rate. By default, $\alpha$ is chosen to be $10$. Then, our optimization goal for the attacking model in adversarial training is to maximize the Euclidean distance between the rescaled softmax output $\hat{\mathbf{h}}$ and the one-hot encoded ground truth vector $\mathbf{y}$: $L_2 = ||\hat{\mathbf{h}}-\mathbf{y}||_2$. We further perturb the target loss by adding a Gaussian noise $\mathcal{N}(0, \sigma)$ with magnitude $\sigma$. Such perturbed attack loss is then:
$ L_\mathcal{N} = ||\hat{\mathbf{h}}+\mathcal{N}(0, \sigma)-\mathbf{y}||_2 $.

An adaptive searching rule is designed to downscale the perturbation by a factor $\beta$ once the attack succeeds which avoids over-optimized adversarial samples biased far from ID characteristics. Therefore, the attack can maintain a fixed success rate with a minimized perturbation norm.
The \textit{Jitter} loss is then:
\begin{equation}
    L_{Jitter} = 
        \begin{cases}
        \frac{||\hat{\mathbf{h}}+\mathcal{N}(0, \sigma)-\mathbf{y}||_2}{\beta} & \text{if $f_\theta(x_\gamma)=y$},\\
        ||\hat{\mathbf{h}}+\mathcal{N}(0, \sigma)-\mathbf{y}||_2 &  \text{otherwise}.
        \end{cases}
\end{equation}

\subsection{Multi-Geometry Projection (MGP)}
\label{sssec:mmel}

Our backbone network $f_\theta$ incorporates a dual-stream geometry projection to capture diverse latent structures in ID data. Each geometry stream is defined by its specific loss function for joint optimization. In this context, we introduce hypersphere and hyperbolic geometries which are Riemannian manifolds with positive and negative curvature. The curvature serves as an indicator of deviation from the Euclidean space. 
Hyperspherical geometry has shown its effectiveness in OOD detection~\cite{ming2023how}. The hyperbolic space has been used in open-set recognition~\cite{openset} that can model the hierarchical structures found in real-world vision data~\cite{hyperbolic}, as evident in datasets like Imagenet. 

We assume parameter $\theta$ resides on a Riemannian manifold $\mathcal{M}$ with the Riemannian metric tensor $g^{\mathcal{M}}$. The tensor $g^{\mathcal{M}}: \mathcal{T}_\theta\mathcal{M} \times \mathcal{T}_\theta\mathcal{M}$ consists of inner products in its tangent space $\mathcal{T}_\theta\mathcal{M}$. A retraction map $R_\theta$ provides transformations from $\mathcal{M}$ to the tangent space $\mathcal{T}_\theta\mathcal{M}$. 
The tangent space can be regarded as a measure of small deviation $\gamma$ near parameter $\theta$, and the metric $g^{\mathcal{M}}$ smoothly varies across $\theta \in \mathcal{M}$, resulting in the geodesic distance. The deviation $\gamma$ on $\mathcal{T}_\theta\mathcal{M}$ is considered as the perturbation generated for adversarial training (as discussed in $\S$\ref{sssec:adversarial_training}), which will be utilized in Riemannian manifold optimization ($\S$\ref{sssec:optimization}). 

We incorporate the following geometries, each with its own loss metric designs.

\medskip
\noindent\textbf{Hypersphere geometry}: Learning hypersphere geometry involves compactness and disparity loss functions to group data samples onto a hypersphere. These functions ensure that samples from different classes are kept at sufficient distances from each other. 
The hypersphere projection approach initially introduced as CIDER~\cite{ming2023how}, is based on the von Mises-Fisher (vMF) distribution assumption. It is calculated using a unit vector $\mathbf{z_s} \in \mathcal{R}_s^d$ in class $k$ and the class prototype $\boldsymbol{\mu}_k$ as: $p_d(\mathbf{z_s};\boldsymbol{\mu}_k) = \tau \; \exp 
\left( \boldsymbol{\mu}_k\mathbf{z_s}/\tau
\right),$
where $\tau$ is a temperature parameter.
The probability of the embedding $\mathbf{z_s}$ assigned to class $k$ is:
$\mathcal{P} \left(
    y=k|\mathbf{z_s};\{\boldsymbol{\mu}_k, \tau\}
\right) = \frac{
\exp \left( \boldsymbol{\mu}_k\mathbf{z_s}/\tau
\right)}{\sum_{j=1}^{K}{
\exp \left( \boldsymbol{\mu}_j\mathbf{z_s}/\tau
\right)}}.$
We derive the {\em compactness loss} by taking negative log-likelihood, which compels the projected samples to stay near the class prototypes.
\begin{equation}
    \mathcal{L}_{com} = -\frac{1}{N}\log\frac{\exp(\boldsymbol{\mu}_k\mathbf{z_s}/\tau)}{\sum_{j=1}^{K}{\exp(\boldsymbol{\mu}_j\mathbf{z_s}/\tau)}}.
\label{eq:comp_loss}
\end{equation}
The {\em disparity loss} penalizes the class prototypes that are too close to each other:
\begin{equation}
    \mathcal{L}_{dis} = -\frac{1}{K}\sum_{i=1}^{K}\log\frac{1}{K-1}\sum_{j=1}^{K}\mathbf{1}_{ji}
    \exp{ 
      \left( \boldsymbol{\mu}_i\boldsymbol{\mu}_j/\tau
      \right)
    },
\label{eq:disparity_loss}
\end{equation}  
where $\mathbf{1}_{ji}$ is indication function, 
$\mathbf{1}_{ji} =
    \begin{cases}
      1 & \text{if $j\neq i$},\\
      0 & \text{otherwise.}
    \end{cases}$
The {\em hypersphere loss} function is $\mathcal{L}_{sph} = \mathcal{L}_{com} + \mathcal{L}_{dis}$, which imposes constraints on ID intra-class compactness and inter-class disparity on the hypersphere. Meanwhile, OOD data are more likely to be separated farther from ID prototypes.

\medskip
\noindent\textbf{Hyperbolic geometry}: A hyperbolic space $H^d$ consists of $d$-dimensional Riemannian manifolds with constant negative curvature~\cite{khrulkov2020hyperbolic}. 
An isomorphic hyperbolic transformation, Poincaré Ball ($\mathcal{D}^d_c, g^\mathcal{D}$), defines a manifold $\mathcal{D}^d=\{\mathbf{u}\in \mathbb{R}^d: c||\mathbf{u}||<1\}$ equipped with the Riemannian metric $g^\mathcal{D}(\mathbf{u}) = (\lambda^c_\mathbf{u})^2g^E = (\frac{2}{1-c||\mathbf{u}||^2})^2\mathbf{I}$, where $\lambda = \frac{2}{1-c||\mathbf{u}||^2}$ is a conformal factor with curvature $c$, and $g^E=\mathbf{I}$ is an Euclidean metric tensor. The manifold operates on Mobius gyrovector space with Mobius addition $\oplus_c$ and scalar multiplication $\otimes_c$ (referring to appendix A.1).

The pairwise geodesic distance is in the following form for two points $\mathbf{u}$ and $\mathbf{v}$: $D(\mathbf{u}, \mathbf{v}) = \frac{2}{\sqrt{c}}\arctanh \left(\sqrt{c}||-\mathbf{u} \oplus_c \mathbf{v}|| \right).$
%
Utilizing the operations of the hyperbolic space, we project the latent embedding with a hyperbolic head to derive the embedding $\mathbf{u}$ on the Poincaré ball. Considering an augmented set $\mathcal{A}$ from $\mathcal{X}$ to form a full set $\mathcal{I}=\mathcal{A}\cup \mathcal{X}$, the supervised contrastive loss is calculated on the positive sample $p(i)$ of the $i\in \mathcal{I}$ in contrast to other augmented samples $a\in \mathcal{A}$. 
The supervised hyperbolic contrastive loss can thus be formulated as $\mathcal{L}_{hypb} = $
\begin{equation}
-\sum_{i\in \mathcal{I}}\frac{1}{|P(i)|}\sum_{p\in P(i)}\log\frac{\exp
\left( -D(\mathbf{z}_i, \mathbf{z_h}_p)/\tau \right)}{\sum_{a\in \mathcal{A}}{\exp
\left( -D(\mathbf{z_h}_i, \mathbf{z_h}_a)/\tau \right)}}.
\nonumber    
\end{equation}
The final loss is the combination of the hypersphere and hyperbolic losses, along with a cross-entropy loss $\mathcal{L}_{ce}$ to optimize for ID classification accuracy: $\mathcal{L} = \mathcal{L}_{sph} + \mathcal{L}_{hypb} + \mathcal{L}_{ce}.$

\subsection{Riemannian Sharpness-aware Minimization}
\label{sssec:optimization}

\noindent\textbf{Theoretical statement}:
The practice of adversarial training serves as a countermeasure to defense against adversarial attacks. However, previous studies have uncovered challenges related to model convergence, attributed to the intricate loss landscape shaped by adversarial samples during training steps~\cite{yamada2021adversarial,cheng2022adversarial}. Next, we delve into the theoretical underpinnings of this phenomenon.

The weight of a $j$-th layer in a neural network $f_\theta$ can be expressed as: $W^T = W^T_j\Sigma_{j-1}\dots W^T_2\Sigma_2 W^T_1\Sigma_1$, which generetes gradient $g_W = \frac{\partial}{\partial W}L(x, y; f_\theta)$. Introducing adversarial samples $x^\ast = x+\gamma$, the corresponding gradient becomes $g_W^\ast= \frac{\partial}{\partial W}L(x+\gamma, y; f_\theta)$. Considering  $\Delta g_W = g_W^\ast-g_W$ as the additional gradient values resulting from adversarial training, the full expression for gradient $\Delta g_W$ is as follows:
\begin{equation}
    \Delta g_W = \frac{\partial}{\partial W}L(x, y; f_\theta) - \frac{\partial}{\partial W}L(x+\gamma, y; f_\theta), 
\end{equation}
where $\gamma$ can be regularized by a $l_2$ or $l_\infty$ norm. This adversarial perturbation $\gamma$ is generated by $m$ steps of attacks with each step size $\alpha$, where $m$ should be large and $\alpha$ is small. 

we analyze the case of binary classification, where the multi-class classification tasks can be simplified by focusing on the difference between the prediction for the targeted class $z'_1$ and the second highest probable class $z'_2$. Misclassification occurs when the probability of the second class surpasses that of the targeted class, and this difference is denoted as $z = z'_1 - z'_2 \in\mathbb{R}$. The effect of introducing adversarial samples brings in a change of the gradient $\tilde{g}_x = \frac{\partial z(x)}{\partial x}$. The additional gradient for updating the model with a learning rate $\eta$ in adversarial training is expressed as $\Delta \tilde{g}_x = -\eta\Delta g_W\tilde{g}_h$. Here, $\tilde{g}_h=\frac{\partial z(x)}{\partial h}$ indicates the gradient of the network output $z(x)$ concerning the latent layer $h$.   

Based on a lamma described in~\cite{cheng2022adversarial}, we know that the following relations hold with $\mathcal{A}=m\alpha H_z||\tilde{g}_x||^2 \in \mathbb{R}$:
\begin{equation}
    H_x\Delta \tilde{g}_W = (e^\mathcal{A}-1)H_x x \tilde{g}_h^T - \frac{1}{H_z ||\tilde{g}_x||^2}(e^{2\mathcal{A}}-e^{\mathcal{A}})H_x g_x g_h^T.
\end{equation}
The Hessian matrix is $H_h=\frac{\partial^2}{{\partial h\partial h}^T}L(x+\gamma, y; f_\theta)$ which can be rewritten as $H_h=H_z \tilde{g}_h\tilde{g}_h^T$ and $H_x=H_z \tilde{g}_x\tilde{g}_x^T$. 

Therefore, we can assess the significance of this change $\Delta \tilde{g}_x$ along the direction of $\tilde{g}_x$:
\begin{align}
&\tilde{g}_x^T\Delta \tilde{g}_x 
= -\eta\tilde{g}_x^T\Delta g_W\tilde{g}_h \\
&= (e^\mathcal{A}-1)\tilde{g}_x^T\Delta\tilde{g}_x^0 - \frac{\eta g_z^2||\tilde{g}_h||^2}{H_z}(e^{2\mathcal{A}}-e^{\mathcal{A}}), 
\label{eq:x_gradient}
\end{align}
where $\Delta\tilde{g}_x^0=-\eta g_W\tilde{g}_h$. Meanwhile, the significance measuring for the adversarial training along the direction of $\tilde{g}_x$ is as follows:
\begin{align}
&\tilde{g}_x^T\Delta \tilde{g}_x^\ast 
= -\eta\tilde{g}_x^T\Delta g_W^\ast\tilde{g}_h \\
&= e^\mathcal{A}\tilde{g}_x^T\Delta\tilde{g}_x^0 - \frac{\eta g_z^2 e^{2\mathcal{A}}-e^{\mathcal{A}}}{H_z} ||\tilde{g}_h||^2.
\label{eq:adv_gradient}
\end{align}
This design of adversarial training expects the gradient $g_x$ with $g_x^T \Delta \tilde{g}_x < 0$. However, this assumption might not be held as the second term of equation~\eqref{eq:x_gradient} and equation~\eqref{eq:adv_gradient} can be negative owing to $H_z>0$. A few unconfident samples generate large values for $H_z$ and large gradient values $||\tilde{g}_x||$. The phenomenon leads to difficulties in model convergence during adversarial training.

\noindent\textbf{Riemannian optimization strategy}:
Learning complex latent geometries may magnify undesirable peaks in the loss minimization process. Inspired by SAM~\cite{foret2021sharpnessaware} which was originally crafted for model generalization, we employ an improved approach for Riemannian manifolds~\cite{yun2023riemannian,truong2023rsam} tailored to our multi-geometry network. The consideration of multiple geometries in the network represents various manifolds that might not consistently converge in the same gradient direction. The recent work~\cite{ming2023how} only accounts for a single hypersphere geometry, which limits the ability to represent the OOD space. In our scenario, we aim to utilize the Riemannian manifold optimization strategy to strengthen multiple geometries. 

Given a loss function $\mathcal{L}(\theta)$ with model parameter $\theta\in \mathcal{M}$ and retraction map $R_\theta$, the {\em manifold sharpness} is defined as
$
\mathcal{L}_S = \underset{||\delta||^2_\theta\leq \rho}{\operatorname{max}}\mathcal{L}
\left(
R_\theta(\delta)
\right) - \mathcal{L}(\theta),
$
where $\delta$ is a projected perturbation in the tangent space $\mathcal{T}_\theta \mathcal{M}$ of the manifold $\mathcal{M}$. The minimization of $\underset{\theta\in \mathcal{M}}{\operatorname{min}} \mathcal{L}_S$ reduces loss sharpness. 

We simplify the first term in $\mathcal{L}_S$ using Taylor expansion to approximate perturbed loss in the maximization process:  
$
    \mathcal{L}(R_\theta(\delta)) \approx \mathcal{L}(\theta) + 
        \langle \nabla_\theta \mathcal{L}(\theta), \delta 
        \rangle_\theta,
$
where $\nabla_\theta$ denotes the Riemannian gradient. A closed-form solution for $\mathcal{L}_S$ is picking $\delta$ equal to the Riemannian gradient within the upper bound $\rho$. The optimal perturbation is then
$
    \delta^\ast = \rho \frac{\nabla_\theta(\mathcal{L}(\theta))}{||\nabla_\theta( \mathcal{L}(\theta)||_\theta}.
$
We project $\delta^\ast$ onto the tangent space via $R_\theta$ and derive the optimal parameter $\theta^\ast=R_\theta \left( \delta^\ast \right)$. The network parameter in the next training iteration $\theta'$ can be updated using Riemannian gradient descent as:
$    \theta' = R_\theta
    \left( -\eta \cdot \nabla_\theta (\mathcal{L}(\theta^\ast)) \right),$
where $\eta$ is the learning rate.
During the adversarial training described in $\S$\ref{sssec:adversarial_training}, the sharpness $\mathcal{L}_S$ on the loss landscape would unexpectedly increase. Our solution is introducing RSAM, which can regularize the network to increase convergence quality.

\subsection{OOD Scoring Function}

With a trained network $f$ in the MGP framework, we extract the penultimate layer output as an L2 normalized embedding $\mathbf{z}$ of the sample $\mathbf{x}$ to compute its OOD score. 
To distinguish OOD from ID samples, we calculate the embedding distance between each input sample and the training ID samples and specify the $k^{th}$ nearest neighbor as a reference embedding $\mathbf{z}_k$. 
The OOD score is based on the L2 distance, $S(\mathbf{z}) = ||\mathbf{z}-\mathbf{z}_k||_2$ that determines the detection of an OOD sample via a cutoff threshold $\lambda$.

\section{Experiments}
\label{sec:experiment}

\noindent
{\bf Dataset:}
Our OOD detection experiments are categorized into results for approaches with and without defense. For OOD detection without adversarial defense, we use CIFAR-10 and CIFAR-100~\cite{krizhevsky2009learning} as the ID dataset, and evaluate the performance on six other datasets that are treated as OOD: Tiny-ImageNet~\cite{le2015tiny}, Place365~\cite{7968387}, LSUN~\cite{yu2015lsun}, LSUN-Resize~\cite{yu2015lsun}, iSUN~\cite{xu2015turkergaze}, and Textures~\cite{6909856}. 
For the compared OOD detection with adversarial defense, ATOM~\cite{chen2021atom} and ATD~\cite{azizmalayeri2022your} requires Food-101~\cite{bossard2014food} dataset for additional outlier data training and SVHN~\cite{netzer2011reading} dataset for validation.

%

\noindent
{\bf Evaluation metric:}
(1) FPR$_{95}$: False positive rate at true positive rate 95\% in the Receiver Operating Characteristic (ROC) analysis.
(2) AUC: Area under the ROC curve.

\noindent
{\bf Attack setup:}
%
We investigate a set of attacks including PGD~\cite{madry2018towards}, FGSM~\cite{goodfellow2015explaining}, FAB~\cite{croce2020minimally}, Jitter~\cite{schwinn2023exploring}, and Carlini and Wagner Attack (CW)~\cite{carlini2017towards}, which are implemented using the TorchAttacks toolbox~\cite{kim2020torchattacks}. 
The attacks are constrained with perturbation bound $\epsilon=\frac{8}{255}$ and step size $\frac{2}{255}$ for 10-step iterations.

\noindent
{\bf Model Configurations:}
Our CIFAR-10 evaluation uses a ResNet-18 backbone network and CIFAR-100 uses ResNet-34. The base optimizer is stochastic gradient descent (SGD) with momentum 0.9, weight decay \text{$10^{-4}$}, and an initial learning rate of 0.5. This optimizer is regularized by RSAM in $\S$\ref{sssec:optimization}. The model undergoes training for 500 epochs with a batch size of 512. We specify the intermediate layer with 128 dimensions. 
The curvature $c$ of hyperbolic geometry is set to be $0.01$.

For the stability of learning, we adopt the feature clipping technique that is empirically found useful for better convergence~\cite{9880306} and to avoid the gradient vanishing of complex manifold learning. An Euclidean space sample point $\mathbf{x}$ is truncated as the clipped feature $\mathbf{x'}=\min\{1, \frac{r}{||\mathbf{x}||}\}\cdot \mathbf{x}$ with the effective radius $r$ of the Poincaré ball. This process regularizes the points sitting overly close to the ball boundary. 


\begin{table*}[t]
\addtolength{\tabcolsep}{1pt}
\caption{Evaluation of OOD detection methods {\bf without adversarial training} and {\bf with adversarial training} using \textit{CIFAR-10} as the ID dataset. We report the average FPR$_{95}$ and AUC scores across the six OOD datasets. Apart from the ``Clean'' setting, ``Adversarial'' conditions denote the further average results over five attacks (PGD, Jitter, FAB, FGSM, and CW). The complete results of each OOD dataset are presented in the appendix.
}
\label{tab:sota_cifar10}
\centerline{
\setlength{\tabcolsep}{1.8mm}
\begin{tabular}{l|cc|cc|cc|cc|cc|cc|cc}
\toprule
\multicolumn{15}{c}{Without Adversarial Training}  \\
\midrule
Condition & \multicolumn{2}{c}{Clean}  & \multicolumn{2}{c}{PGD}             & \multicolumn{2}{c}{Jitter}    & \multicolumn{2}{c}{FAB} & \multicolumn{2}{c}{FGSM} & \multicolumn{2}{c}{CW}  & \multicolumn{2}{c}{Average} \\
\midrule
Metric & FPR$_{95}$$\downarrow$   & AUC   & FPR$_{95}$$\downarrow$            & AUC            & FPR$_{95}$$\downarrow$           & AUC            & FPR$_{95}$$\downarrow$            & AUC            & FPR$_{95}$$\downarrow$            & AUC     & FPR$_{95}$$\downarrow$            & AUC  & FPR$_{95}$$\downarrow$ & AUC \\
\midrule
Entropy    & 32.18 & 91.59  & 95.14  & 48.00 & 90.46  & 59.14   &  95.14  & 48.00    &  93.75  & 52.83 & 96.00 &	47.16 & 82.11 & 57.79 \\
ViM      &  29.17 & 92.98  & 91.65  & 60.59 &  81.68  & 71.03  &  91.65  & 60.59    &    87.61  & 65.78 & 92.20 &	60.14 & 78.99 & 68.52\\
Mahalanobis   & 17.46 & 96.84  & 75.03 & 73.58 &  70.96 & 74.83 &  75.04 & 73.57    & 74.59         & 73.55 &    76.23 &	72.53 & 64.89 & 77.48 \\
ODIN     & 42.51 & 91.12 & 92.93  & 55.22 & 89.71  & 63.30  &  92.93  & 55.22 &  91.38  & 61.47 &    94.68 & 53.98 & 84.02 & 63.39 \\
GODIN    & 18.72 & 96.10 & 70.87 & 83.81 &  60.59 & 84.76 & 71.08 & 83.75 &    70.33 & 83.23 &   70.68 & 83.95  & 60.38 & 85.93 \\
ASH     & 27.53     & 94.08     & 71.31     & 78.45     & 68.54     & 79.58 &  81.79 & 71.12  & 87.07   & 67.86   & 81.66   & 70.90  &  69.65  & 77.00  \\
KNN+    & \textbf{18.06}  & \textbf{96.59} & 77.52   & 78.5    & 77.41       & 78.83   & 76.32      & 75.14   & 77.26      & 78.81   & 76.26       & 79.65   & 67.14 &  81.25  \\
SSD    & 33.08   & 94.87   & 79.72   & 65.01   & 78.28       & 68.3    & 80.03      & 65.13   & 64.13      & 78.88   & 79.39       & 65.08   & 69.11    & 72.88    \\
CIDER-KNN      & 52.20    & 88.41   & 66.73   & 76.92   & 66.48       & 78.73   & 66.35      & 76.95   & 72.86      & 72.50    & 66.45       & 76.83   & 65.18    & 78.39   \\
CIDER-Maha   & 51.19   & 88.91   & 58.95   & 83.47   & 54.29       & 87.43   & 59.32      & 83.49   & 49.96      & 88.42   & 59.41       & 83.51   & 55.52    & 85.87 \\
MGP-KNN     &  21.60    & 96.11   & 82.49   & 72.02   & 70.64       & 81.71   & 78.27      & 77.95   & 78.27      & 77.95   & 83.22       & 71.2    & 69.08    & 79.49 \\
MGP-Maha   & 29.98   & 95.36   & \textbf{55.39}   & \textbf{87.81}   & \textbf{50.43}       & \textbf{90.52}   & \textbf{55.49}      & \textbf{87.93}   & \textbf{38.91}      & \textbf{91.82}   & \textbf{54.28}       & \textbf{87.92}   & \textbf{47.41}    & \textbf{90.23}   \\
\midrule\midrule
\multicolumn{15}{c}{With Adversarial Training}  \\
\midrule
Condition & \multicolumn{2}{c}{Clean}  & \multicolumn{2}{c}{PGD}             & \multicolumn{2}{c}{Jitter}    & \multicolumn{2}{c}{FAB} & \multicolumn{2}{c}{FGSM} & \multicolumn{2}{c}{CW}  & \multicolumn{2}{c}{Average} \\
\midrule
Metric & FPR$_{95}$$\downarrow$   & AUC   & FPR$_{95}$$\downarrow$            & AUC            & FPR$_{95}$$\downarrow$           & AUC            & FPR$_{95}$$\downarrow$            & AUC            & FPR$_{95}$$\downarrow$            & AUC     & FPR$_{95}$$\downarrow$            & AUC  & FPR$_{95}$$\downarrow$ & AUC \\
\midrule
ATOM    &  23.57  & 88.22  &  66.14   &  72.56   &    55.60     &     79.17     &    63.20     &   79.93      &     55.08     &  80.36  &  73.19  &  74.24    & 56.13     & 79.08  \\
ATD    &  27.46   & 94.15   & 57.04    & 79.63    & 43.61    & 87.60    & 33.09  & 92.38  & 37.56  & 90.70  & 56.78  & 79.69  & 42.59 & 87.36 \\
SaGD  & \textbf{22.46}     & \textbf{95.77}     & \textbf{28.69}     & \textbf{94.70}     & \textbf{26.43}     & \textbf{95.05}   & \textbf{28.99}   & \textbf{94.68} & \textbf{37.19} & \textbf{92.99}   & \textbf{22.32} & \textbf{95.80} & \textbf{27.68} & \textbf{94.83} \\
\bottomrule
\end{tabular}
}
\end{table*}

\begin{table*}[t]
\addtolength{\tabcolsep}{1pt}
\caption{Evaluation of OOD detection methods {\bf without adversarial training} and {\bf with adversarial training} using \textit{CIFAR-100} as the ID dataset.
}
\label{tab:sota_cifar100}
\centerline{
\setlength{\tabcolsep}{1.8mm}
\begin{tabular}{l|cc|cc|cc|cc|cc|cc|cc}
\toprule
\multicolumn{15}{c}{Without Adversarial Training}  \\
\midrule
Condition & \multicolumn{2}{c}{Clean}  & \multicolumn{2}{c}{PGD}             & \multicolumn{2}{c}{Jitter}    & \multicolumn{2}{c}{FAB} & \multicolumn{2}{c}{FGSM} & \multicolumn{2}{c}{CW}  & \multicolumn{2}{c}{Average} \\
\midrule
Metric & FPR$_{95}$$\downarrow$   & AUC   & FPR$_{95}$$\downarrow$            & AUC            & FPR$_{95}$$\downarrow$           & AUC            & FPR$_{95}$$\downarrow$            & AUC            & FPR$_{95}$$\downarrow$            & AUC     & FPR$_{95}$$\downarrow$            & AUC  & FPR$_{95}$$\downarrow$ & AUC \\
\midrule
Entropy    & 88.62 & 60.42 & 92.36  & 53.65 &  87.96  & 61.67          & 79.30        & 62.14        & 93.03        & 49.89 & 92.11 & 53.88  & 88.90 & 56.94 \\
ViM     & 75.94 & 73.34 & 83.55  & 63.17 & 80.16  & 65.52     & 74.00        & 69.21        & 84.34        & 58.72   & 80.10 & 68.91 & 79.68 & 66.48 \\
Mahalanobis  & 72.21 & 74.22 & 81.38 & 63.00 & 81.53 & 63.07 & 77.23 & 66.09        & 82.79        & 59.45  & 75.42 & 69.63 & 78.43 & 65.91 \\
ODIN    & 81.57 & 68.05 & 89.86  & 58.54 & 84.56 & 63.94  & 76.88        & 62.07        & 91.18        & 54.42   & 89.96 & 58.33 & 85.67 & 60.89 \\
GODIN    & 74.58 & 80.88  & 90.19 & 67.37  & 90.45 & 73.03 & 94.66 & 66.41 & 95.33        & 64.85 & 90.39 & 67.39 & 89.27 & 69.99 \\
ASH    & 59.04 & 84.44 & 73.74 & 62.56  &   70.37  & 77.93 & 78.18 &	70.14 & 85.89 & 65.01  & 75.24 & 75.72 & 73.69 & 72.63 \\
KNN+    & 65.47 & 85.07 & 95.59 & 51.07 & 95.39 & 51.38 & 95.44 & 51.61 & 95.55 & 50.53 & 95.38 & 51.58 & 90.47 & 56.87 \\
SSD    & 70.98 & 84.94 & 95.16 & 46.80 & 94.51 & 49.60 & 95.28 & 46.75 & 92.31 & 52.54 & 95.10 & 46.60 & 90.56 & 54.53 \\
CIDER-KNN     & 65.99  & 83.44  & 75.69   & \textbf{82.95}   & 66.59 & 82.07 & \textbf{74.70}           & \textbf{83.08} & 78.24 & 77.43 & 76.10        & 83.06 & 72.88    & \textbf{82.01}     \\
CIDER-Maha   & 67.28   & 84.36 & \textbf{75.47} & 75.85   & \textbf{63.36} & \textbf{83.94} & 74.83 & 75.76 & \textbf{53.26} & \textbf{84.38} & 76.17 & 75.91 & \textbf{68.40}    & 80.03      \\
MGP-KNN    & \textbf{57.89}   & \textbf{85.26}   & 80.23   & 76.17   & 81.41 & 74.63 & 79.80 & 76.28 & 78.74 & 69.99 & \textbf{60.05} & \textbf{82.79} & 73.02 & 77.52   \\
MGP-Maha   & 66.47   & 83.19   & 79.33   & 65.81   & 74.12 & 74.96 & 81.32 & 62.72         & 64.99 & 75.98 & 76.60 & 73.09 & 73.80 & 72.62  \\ 
\midrule\midrule
\multicolumn{15}{c}{With Adversarial Training}  \\
\midrule
Condition & \multicolumn{2}{c}{Clean}  & \multicolumn{2}{c}{PGD}             & \multicolumn{2}{c}{Jitter}    & \multicolumn{2}{c}{FAB} & \multicolumn{2}{c}{FGSM} & \multicolumn{2}{c}{CW}  & \multicolumn{2}{c}{Average} \\
\midrule
Metric & FPR$_{95}$$\downarrow$   & AUC   & FPR$_{95}$$\downarrow$            & AUC            & FPR$_{95}$$\downarrow$           & AUC            & FPR$_{95}$$\downarrow$            & AUC            & FPR$_{95}$$\downarrow$            & AUC     & FPR$_{95}$$\downarrow$            & AUC  & FPR$_{95}$$\downarrow$ & AUC \\
\midrule
ATOM    & 72.81 &  79.31 &   79.35  &  78.48   &     76.87    &    80.53      &    74.98     &   82.71      &     79.25     &  81.01  &  85.66  & 78.68    & 76.82 &  80.12 \\
ATD     & 63.04 &  82.90  &  73.79  &  66.18  & 64.90       & 80.07        & 63.04       & 82.90        & 70.38        & 76.37   & 70.30 & 76.06   & 67.58 & 77.41 \\
SaGD   & \textbf{50.89}     & \textbf{87.26}    & \textbf{48.31}     & \textbf{88.13}   & \textbf{47.63}     & \textbf{88.15}     & \textbf{49.08} & \textbf{87.88}  & \textbf{52.51}   & \textbf{86.78}  & \textbf{50.81} & \textbf{87.32} & \textbf{49.87}   & \textbf{87.59}   \\
\bottomrule
\end{tabular}
}
\end{table*}
\subsection{Evaluation of out-of-distribution accuracy}
\label{sssec:exp_ood_accuracy}

We report the averaged OOD detection results over the six OOD datasets. The full results for each dataset under different attacks are reported in the appendix.

\noindent\textbf{Without Defense}:
The upper part of Table~\ref{tab:sota_cifar10} and Table~\ref{tab:sota_cifar100} showcase popular OOD detection methods under adversarial attacks using CIFAR-10 and CIFAR-100 as the ID dataset, respectively.
We report post-processing-based OOD detection baselines including Entropy~\cite{chan2021entropy}, 
Mahalanobis~\cite{lee2018simple}, MaxSoftmax~\cite{hendrycks2016baseline}, Energy~\cite{liu2020energy}, MaxLogits~\cite{hendrycks2022scaling}, KLMatching~\cite{hendrycks2022scaling}, ODIN~\cite{liang2018enhancing}, ViM~\cite{wang2022vim}, GODIN~\cite{hsu2020generalized}, and ASH~\cite{djurisic2022extremely}.
We compare with embedding-based methods without adversarial training including SSD~\cite{sehwag2020ssd}, KNN+~\cite{sun2022out}, and CIDER~\cite{ming2023how}, and MGP approach ($\S$\ref{sssec:mmel}). We consider both KNN~\cite{sun2022out} and Mahalanobis~\cite{lee2018simple} as scoring functions for CIDER and MGP, which are denoted in the form of `detector-function' in Table~\ref{tab:sota_cifar10} and Table~\ref{tab:sota_cifar100}. 
These OOD approaches are not designed to defend against malicious attacks. Thus, the experiment can reflect performance degradation under attacks. 

In differentiating the CIFAR-10 dataset from the other OOD dataset shown in Table~\ref{tab:sota_cifar10}, KNN+ attains notable FPR$_{95}$ and AUC in clean conditions while suffering significant degradation in different adversarial conditions. The resulting 67.14\% FPR$_{95}$ and 81.25 AUC is far from satisfactory. In contrast, MGP-Maha outperforms other methods with 47.47\% averaged FPR$_{95}$ and 90.23\% averaged AUC over six OOD datasets, suggesting the robustness of multi-geometry embedding learning relative to the hypersphere geometry shown as CIDER-Maha. Post-training processing approaches such as Entropy, ODIN, and ASH are sensitive to adversarial attacks and thus lead to high FPR$_{95}$ results. 
In general, Mahalanobis scores (Maha)~\cite{lee2018simple} stands out as a remarkable scoring function for lower FPR$_{95}$ while the KNN scoring function is still comparable.

In differentiating the CIFAR-100 dataset from the other OOD dataset shown in Table~\ref{tab:sota_cifar100}, the difference between CIDER and MGP is relatively minor. CIDER-Maha reaches 68.40\% averaged FPR$_{95}$ and CIDER-KNN attains 82.01\% averaged AUC over six OOD datasets. The averaged FPR$_{95}$ and AUC are 73.02\% and 77.52\% for MGP-KNN showing a gap in comparing the clean condition with 57.89\% FPR$_{95}$ and 85.26\% AUC. These OOD detection algorithms face non-ignorable risks in adversarial situations.

\noindent\textbf{With Defense}: 
In Table~\ref{tab:sota_cifar10} and Table~\ref{tab:sota_cifar100}, we compare our proposed SaGD approach to the SToA adversarial defense methods, ATOM~\cite{chen2021atom} and ATD~\cite{azizmalayeri2022your} using the CIFAR-10 and CIFAR-100 ID datasets. 
SaGD achieves notable performance with average FPR$_{95}$ of 27.68\% and AUC of 94.83\% using CIFAR-10 as ID data. For CIFAR-100 as the ID data, SaGD attains an average FPR$_{95}$ of 50.03\% and an AUC of 87.53\%, outperforming ATD significantly.
The confidence-based algorithms such as CCU and ACET are not resilient to adversarial conditions, with average FPR$_{95}$ values over 60\% and 80\% for the CIFAR-10 and CIFAR-100 datasets. Although ATOM obtains a 23.57\% average FPR$_{95}$ in clean OOD detection using the CIFAR-10 ID data, the adversarial results are still inferior to SaGD and ATD. 
%
SaGD demonstrates substantial superiority over ATD by at least 17\% on the CIFAR-100 ID dataset. For the clean set without attacks, ATD achieves a relatively close AUC to SaGD on the CIFAR-10 dataset but falls short by 4.64\%. Notably, the difference in FPR$_{95}$ is substantial, with SaGD achieving 5.00\% and 12.15\% lower FPR$_{95}$ than ATD on CIFAR-10 and CIFAR-100 datasets, respectively.   
The more difficult OOD detection conditions of CIFAR-100 reveal even more pronounced advantages of using SaGD.

Another advantage of our SaGD is its ability to circumvent the need for additional outlier datasets, a requirement in ATD and ATOM for performing outlier exposure.

\begin{table*}[t]
\caption{Ablation study results on CIFAR-10. The upper part presents the ablation of modules in SaGD including MGP/CIDER network, Jitter adversarial training, and RSAM optimization. Our proposed SaGD is located in the last row of the upper table (MGP+RSAM+Jitter). The lower part is about replacing Jitter with other perturbations for adversarial training.
}
\label{tab:ablation}
\centerline{
\setlength{\tabcolsep}{1.9mm}
\footnotesize
\begin{tabular}{c|c|cc|cc|cc|cc|cc|cc|cc}
\toprule
\multicolumn{2}{c}{CIDER} & \multicolumn{2}{c}{Clean}  & \multicolumn{2}{c}{PGD}             & \multicolumn{2}{c}{Jitter}    & \multicolumn{2}{c}{FAB} & \multicolumn{2}{c}{FGSM} & \multicolumn{2}{c}{CW} & \multicolumn{2}{c}{Average} \\
\midrule
RSAM & AT  &  FPR$_{95}$   & AUC   & FPR$_{95}$            & AUC            & FPR$_{95}$           & AUC            & FPR$_{95}$            & AUC            & FPR$_{95}$            & AUC       & FPR$_{95}$            & AUC  & FPR$_{95}$            & AUC\\
\midrule
\xmark & \xmark & 52.20    & 88.41   & 66.73   & 76.92   & 66.48       & 78.73   & 66.35      & 76.95   & 72.86      & 72.50    & 66.45       & 76.83   & 65.18    & 78.39   \\
\checkmark & \xmark & 62.48     & 86.76     & 77.43     & 71.65     & 73.33     & 76.97     & 69.62 & 75.60  & 75.62 & 77.48 & 64.01 & 85.71 & 70.41 & 79.03  \\
\xmark & Jitter  &  35.23   & 94.12   & 59.18   & 86.67   & 60.57       & 86.5    & 59.68      & 86.52   & 70.55      & 81.59   & 60.03       & 86.6    & 57.54    & 87.00 \\
\checkmark & Jitter & 44.66     & 92.20     & 47.58     & 91.06     & 47.29     & 91.52     & 55.89 & 89.44 & 46.52 & 92.00 & 45.39 & 92.14 & 47.89 & 91.39    \\
\midrule
\multicolumn{2}{c}{MGP}  &  FPR$_{95}$   & AUC   & FPR$_{95}$            & AUC            & FPR$_{95}$           & AUC            & FPR$_{95}$            & AUC            & FPR$_{95}$            & AUC       & FPR$_{95}$            & AUC  & FPR$_{95}$            & AUC\\
\midrule
\xmark & \xmark &  \textbf{21.60}    & \textbf{96.11}   & 82.49   & 72.02   & 70.64       & 81.71   & 78.27      & 77.95   & 78.27      & 77.95   & 83.22       & 71.2    & 69.08    & 79.49 \\
\checkmark & \xmark & 22.84   & 96.01   & 73.22   & 78.65   & 73.22       & 78.65   & 72.89      & 78.89   & 75.55      & 71.07   & 71.89       & 78.94   & 73.35    & 80.37  \\   
\xmark & Jitter & 30.32 & 94.88 & 30.87 & 94.70 & 31.97  & 94.67 & 31.49 & 94.66 & 38.08 & 93.27 & 31.39 & 94.68 &   32.35 & 94.48 \\   
\checkmark & Jitter & 22.46    &  95.77     & \textbf{28.69}     & \textbf{94.70}     & \textbf{26.43}     & \textbf{95.05}   & \textbf{28.99}   & \textbf{94.68} & \textbf{37.19} & \textbf{92.99}   & \textbf{22.32} & \textbf{95.80} & \textbf{27.68} & \textbf{94.83} \\
\midrule\midrule
\multicolumn{2}{c}{SaGD} & \multicolumn{2}{c}{Clean}  & \multicolumn{2}{c}{PGD}             & \multicolumn{2}{c}{Jitter}    & \multicolumn{2}{c}{FAB} & \multicolumn{2}{c}{FGSM} & \multicolumn{2}{c}{CW} & \multicolumn{2}{c}{Average} \\
\midrule
RSAM & AT & FPR$_{95}$   & AUC   & FPR$_{95}$            & AUC            & FPR$_{95}$           & AUC            & FPR$_{95}$            & AUC            & FPR$_{95}$            & AUC       & FPR$_{95}$            & AUC  & FPR$_{95}$            & AUC\\
\midrule
\checkmark & PGD          & 86.64 & 60.31 & 95.62 & 75.42 & 94.54  & 60.95 & 88.25 & 77.74 & \textbf{20.25} & \textbf{95.19} & 86.60  & 60.36 & 78.65 & 71.66 \\
\checkmark & FAB  & 87.00 & 61.31   & 42.72   & 91.13   & 94.23          & 51.71          & 42.92          & 91.15         & 99.46          & 61.23         & 43.79          & 90.96          & 68.35    & 74.58 \\ 
\checkmark & FGSM & 92.67 & 52.23 & 80.51 & 85.46 & 95.07  & 50.34 & 98.46	& 51.10 & 59.20  & 91.20  & 92.71 & 52.24 & 86.44 & 63.76 \\
\checkmark & CW      & 45.44 & 91.67 &  70.24 & 83.88 & 53.60 & 88.42 & 69.69 & 83.77 & 66.71 &  85.35 &   48.77 & 90.83 & 59.08 & 87.32 \\ 
\bottomrule
\end{tabular}
}
\end{table*}

\subsection{Ablation Study}
\label{sssec:exp_ablation}

We conduct an ablation study for related techniques using the CIFAR-10 ID dataset, to elucidate the effects of each module in our SaGD framework. The results are presented in Table~\ref{tab:ablation}.


\noindent\textbf{Ablation study of geometry space, adversarial training, and RSAM:}
The removal of the RSAM optimization module from our proposed SAGD adversely impacts both FPR${95}$ and AUC.
Specifically, MGP-Jitter experiences a decline in average FPR$_{95}$ to 32.35\%, reflecting a 4.67\% reduced margin compared to SaGD. Meanwhile, SaGD maintains a high average AUC of 94.48\%, showing no significant decrease. 
Looking from another perspective, MGP-RSAM, discarding the Jitter adversarial training step from SaGD results in a significant increase of 45.67\% in FPR$_{95}$ and a decrease of 14.46\% in AUC. 
We also simplify the MGP structure as CIDER in SaGD which results in its combination with Jitter and RSAM. Jointly using Jitter and RSAM with CIDER obtains 47.89\% FPR$_{95}$ which shows 22.58\% and 9.65\% improvements over CIDER-RSAM and CIDER-Jitter, respectively. 
Overall, the Jitter adversarial training benefits both CIDER and MGP frameworks. 
These results emphasize the significance of conducting Jitter adversarial training, and the RSAM approach can further facilitate the optimization steps. 

\noindent\textbf{Evaluation on different adversarial training methods:}
Based on the idea of generating adversarial examples for robust model training, we investigate additional adversarial attack approaches for adversarial training. Apart from Jitter, we incorporate PGD, FAB, FGSM, and CW to assess the OOD detection results under these different attacks. The lower part of Table~\ref{tab:ablation} shows the average FPR$_{95}$ and AUC over six OOD testing datasets. Most adversarial attacks lead to substantial performance degradation. For example, PGD and FGSM share similar attack properties, resulting in average FPR$_{95}$ exceeding 80\% for the model subjected to any attacks except FGSM. An intriguing result is observed with PGD, achieving 20.25\% FPR$_{95}$ and a 95.19\% AUC. This suggests that this type of perturbation can generate a notably robust model against the specific type of attack but may not generalize well to others.

\subsection{Testing with Different Adversarial Paramters}
\label{sssec:exp_attack_level}

\begin{figure}
\centerline{
  \includegraphics[width=\linewidth,height=0.4\linewidth]{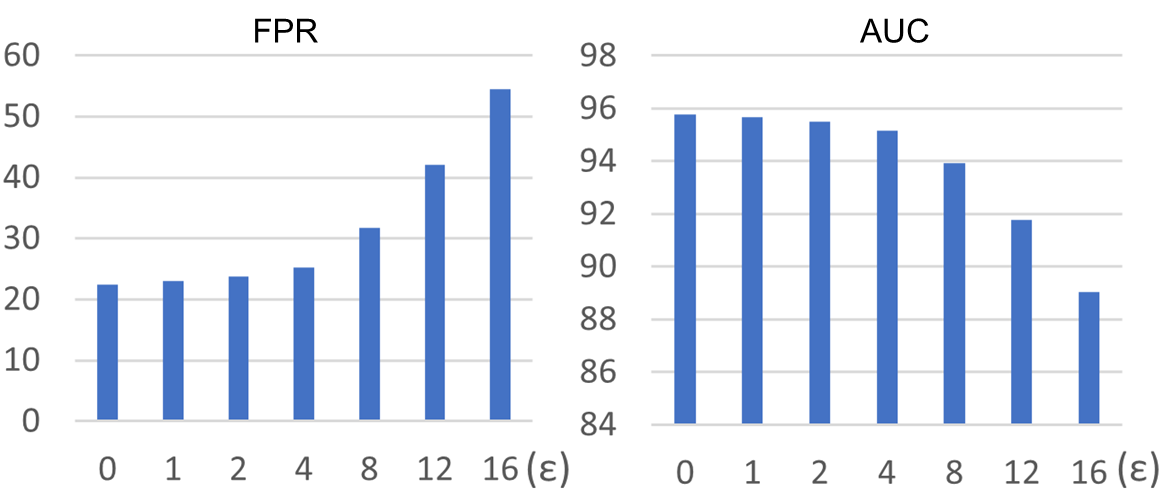}
}
\caption{CIFAR-10 OOD detection results under different PGD attack perturbation intensities ($\epsilon$).}
\label{fig:attack_level}
\end{figure}

\textbf{Perturbation intensity}: We investigate the influence of varying perturbation intensities ($\epsilon$) in the PGD adversarial attack on the SaGD method using the CIFAR-10 dataset. Figure~\ref{fig:attack_level} shows that FPR$_{95}$ is more susceptible to changes, while AUC maintains a consistently high standard as $\epsilon$ increases. Notably, an intense attack with $\epsilon=16$ causes FPR$_{95}$ to double, whereas AUC experiences a 6.72\% decline. These results suggest considering FPR$_{95}$ in the evaluation, an aspect that has been previously overlooked in the literature~\cite{azizmalayeri2022your}.


\textbf{Iterative Attacks}: We include adaptive PGD (APGD) and autoattack~\cite{croce2020reliable} in Table~\ref{tab:add_comparison}. The adaptive PGD is investigated with 100 and 1000 steps and the autoattack is a parameter-free ensemble of multiple attacks. 
Increasing steps for APGD does not significantly affect the performance of ATD and SaGD compared to other types of attacks. The Autoattack obtains similar results with APGD in 1000 steps. SaGD can robustly defend for these different adversarial scenarios. 

\textbf{Inlier AUC and Outlier AUC}:
We analyze adversarial AUC metrics applying to inliers and outliers ($AUC_{In}$ and $AUC_{Out}$) which are also reported in \cite{azizmalayeri2022your}. Our targeted setting performing attacks on ID and OOD data results in lower AUC values than in $AUC_{In}$ and $AUC_{Out}$. SaGD can robustly achieve over 94\% AUC and outperform ATD in the different metrics.



\begin{table*}[t]
\caption{Additional metrics (inlier and outlier AUC) and adaptive attacks on the CIFAR-10 dataset.
}
\label{tab:add_comparison}
\centerline{
\setlength{\tabcolsep}{1.7mm}
\renewcommand{\arraystretch}{0.9}
\begin{tabular}{l|cc|cc|cc|cc|cc}
\toprule
 & \multicolumn{4}{c|}{Average Adversarial}            & \multicolumn{2}{c|}{APGD-100} & \multicolumn{2}{c|}{APGD-1000} & \multicolumn{2}{c}{Autoattack}\\
\midrule
 & FPR$_{95}$            & AUC & AUC$_{In}$ & AUC$_{Out}$ & FPR$_{95}$ & AUC  & FPR$_{95}$ & AUC  & FPR$_{95}$ & AUC  \\
\midrule
ATD     & 42.59     & 87.36  &  88.69 &  89.03  & 43.89 & 88.41  & 44.25 & 85.36 & 47.95 & 83.86 \\
SaGD  & \textbf{27.68}     & \textbf{94.83} & \textbf{95.71} & \textbf{95.86}  & \textbf{28.67} & \textbf{94.18} & \textbf{29.13} & \textbf{94.50} & \textbf{32.18} & \textbf{93.01} \\
\bottomrule
\end{tabular}
}
\vspace{-1mm}
\end{table*}

\subsection{OOD Score Visualization}
\label{ssec:exp_visualization}

\begin{figure}
\centerline{
  \includegraphics[width=\linewidth,height=0.7\linewidth]{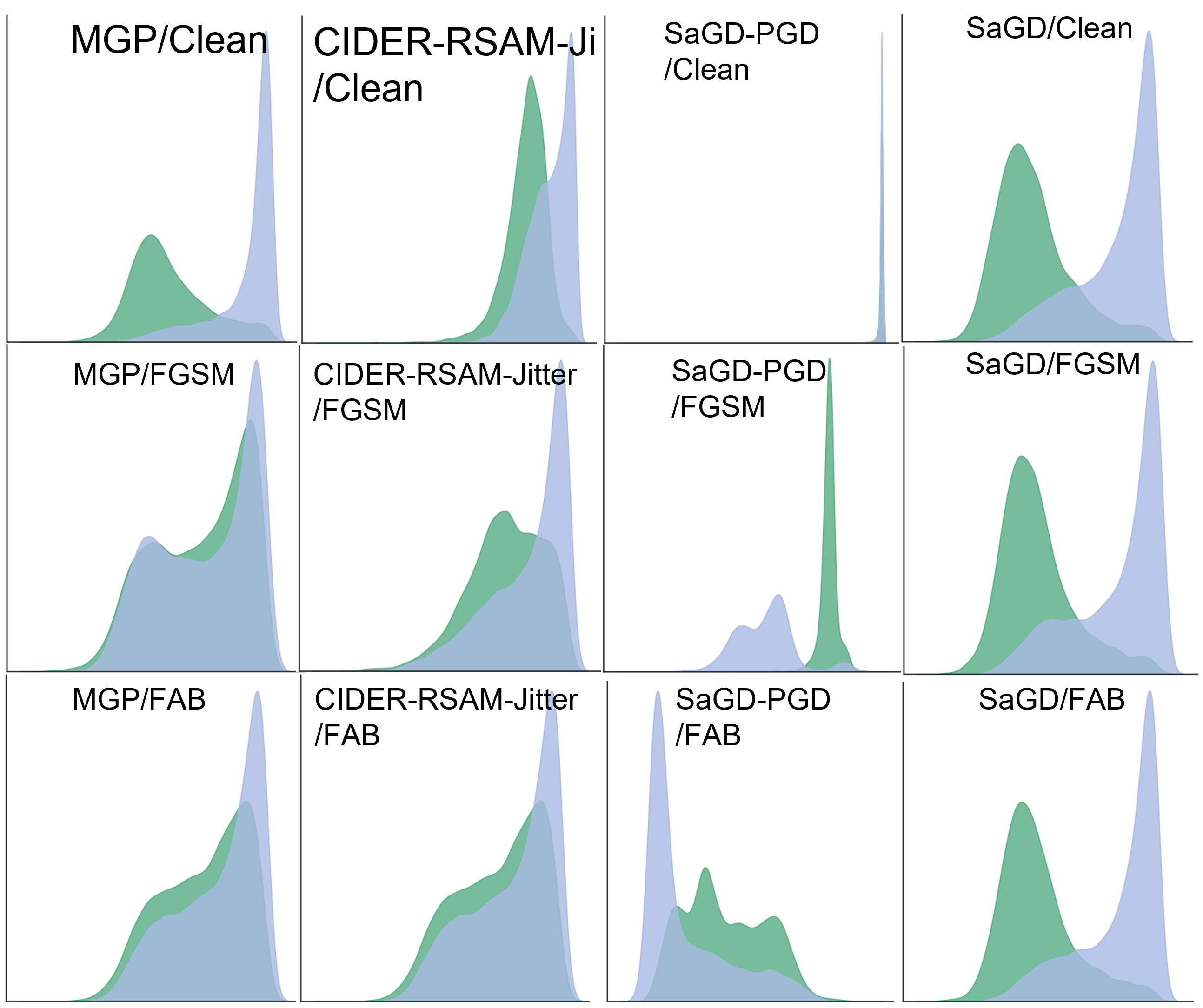} 
  \vspace{-1mm}
}
\caption{ID (blue) and OOD (green) score distribution in the clean condition, FGSM, and FAB adversarial conditions. We denote ``detector/condition'' where the detector can be MGP, CIDER-RSAM-Jitter, SaGD-PGD, or SaGD.}
\label{fig:visualization}
\end{figure}

Figure~\ref{fig:visualization} presents the OOD score histogram distribution between the CIFAR-10 ID testing data and TinyImageNet OOD testing data under clean and adversarial conditions. We demonstrate FGSM and FAB adversarial conditions. Other adversarial results with six OOD datasets are shown in the supplementary file. The ID data are colored in blue and the OOD data are in green. We consider models from the ablation study to further shed light on our proposed technical modules. Specifically, MGP, CIDER-RSAM-Jitter, and SaGD correspond to rows 5, 4, 9, and 8 in Table~\ref{tab:ablation}, respectively. 

In clean conditions, MGP and SaGD distributions look alike, while CIDER-RSAM-Jitter shows a sharper OOD pattern. SaGD-PGD exhibits overlapping distributions between ID and OOD samples, albeit in a narrow area. Under the FGSM attack, MGP and CIDER-RSAM-Jitter distributions collapse significantly, blurring the line between ID and OOD samples. In contrast, SaGD maintains a consistent distribution, preserving a strong discriminative ability even under adversarial conditions. SaGD-PGD produces distinct peaks between ID and OOD distributions against PGD attacks.
Investigating further, under FAB attacks, SaGD-PGD generates multiple peaks in the OOD distribution, confusing it with the long-tailed ID distribution. This highlights the overfitting challenges of adversarial training.
These visualizations illustrate model properties regarding ID and OOD distributions, suggesting the potential of regularizing adversarial optimization across geometry spaces.


\section{Conclusion}
\label{sec:conclusion}

In this paper, we address the robustness issue for out-of-distribution (OOD) detection by investigating various types of adversarial attacks. We propose a novel SaGD framework that leverages the Jitter attack for adversarial training and optimizes the multi-geometry network using RSAM to enhance model convergence. The sharpness minimization strategy mitigates the rugged loss landscape induced by adversarial examples, resulting in improved OOD detection performance under attacks. Our OOD detection experiments encompass two in-distribution (ID) datasets and six OOD datasets tested against five types of attacks. SaGD achieves significantly low FPR$_{95}$ and high AUC on average.   
Our ablation study shows the critical role of Jitter-based adversarial training, highlighting the potential risk of employing popular perturbation approaches like PGD and FGSM. Our analysis shows the importance of using FPR$_{95}$ for evaluation as it tends to be impacted by increased attacks.

\noindent
{\bf Future work} includes the exploration of loss convergence conditions during adversarial geometry learning and improving the generalization of OOD detection capability under various adversarial conditions. We anticipate this work can initiate a novel direction to investigate an in-depth understanding of the relation between geometric loss optimization and robust OOD detection. 



%




\ifCLASSOPTIONcaptionsoff
  \newpage
\fi



\bibliography{IEEEtran/adv_ood}
\bibliographystyle{IEEEtran}

\end{document}